\documentclass[letterpaper, 10 pt, journal, twoside]{IEEEtran}
\usepackage{amsmath} 
\usepackage{amssymb}  
\usepackage{bm}
\usepackage{graphicx}
\usepackage{xcolor}
\usepackage{booktabs}
\usepackage{caption}
\usepackage{subcaption}
\usepackage{algorithmic}
\usepackage[linesnumbered,ruled,vlined]{algorithm2e}
\usepackage{url}

\ifCLASSINFOpdf
\else
\fi
\begin{document}
%
\title{Egocentric Human Trajectory Forecasting with a Wearable Camera and Multi-Modal Fusion}
%
%
%

\author{Jianing Qiu, Lipeng Chen$^\dagger$, Xiao Gu, Frank P.-W. Lo, Ya-Yen Tsai, \\ Jiankai Sun, Jiaqi Liu, and Benny Lo%
\thanks{Manuscript received: February, 24, 2022; Revised May, 20, 2022; Accepted June, 14, 2022.}
\thanks{This paper was recommended for publication by Editor Cesar Cadena Lerma upon evaluation of the Associate Editor and Reviewers' comments.
$^\dagger$ denotes the corresponding author.} 
\thanks{J. Qiu, X. Gu, F. Lo, Y.-Y. Tsai, and B. Lo are with the Hamlyn Centre for Robotic Surgery, Imperial College London, SW7 2AZ, London, UK. J. Qiu, and Y.-Y. Tsai are also with Tencent Robotics X, Shenzhen 518057, China
        {\tt\footnotesize \{jianing.qiu17, xiao.gu17, po.lo15, y.tsai17, benny.lo\}@imperial.ac.uk}}%
\thanks{L. Chen is with Tencent Robotics X, Shenzhen 518057, China
        {\tt\footnotesize lipengchen@tencent.com}}%
\thanks{J. Sun was with Tencent Robotics X, Shenzhen 518057, China and now is with Department of Aeronautics and Astronautics, Stanford University, CA 94305, USA
        {\tt\footnotesize jksun@stanford.edu}}%
\thanks{J. Liu was with Tencent Robotics X, Shenzhen 518057, China and now is with Institute of Medical Robotics, Shanghai Jiao Tong University, Shanghai 200240, China
        {\tt\footnotesize ljq64408@sjtu.edu.cn}}%
\thanks{Digital Object Identifier (DOI): see top of this page.}
}
%
%

\markboth{IEEE Robotics and Automation Letters. Preprint Version. Accepted June, 2022}
{Qiu \MakeLowercase{\textit{et al.}}: Egocentric Human Trajectory Forecasting} 

%



\maketitle

\begin{abstract}
In this paper, we address the problem of forecasting the trajectory of an egocentric camera wearer (ego-person) in crowded spaces. The trajectory forecasting ability learned from the data of different camera wearers walking around in the real world can be transferred to assist visually impaired people in navigation, as well as to instill human navigation behaviours in mobile robots, enabling better human-robot interactions. To this end, a novel egocentric human trajectory forecasting dataset was constructed, containing real trajectories of people navigating in crowded spaces wearing a camera, as well as extracted rich contextual data. We extract and utilize three different modalities to forecast the trajectory of the camera wearer, i.e., his/her past trajectory, the past trajectories of nearby people, and the environment such as the scene semantics or the depth of the scene. A Transformer-based encoder-decoder neural network model, integrated with a novel cascaded cross-attention mechanism that fuses multiple modalities, has been designed to predict the future trajectory of the camera wearer. Extensive experiments have been conducted, with results showing that our model outperforms the state-of-the-art methods in egocentric human trajectory forecasting.   

\end{abstract}


%
\IEEEpeerreviewmaketitle

\section{Introduction}
\label{sec:intro}

\begin{figure}[!t]
\centerline{\includegraphics[width=\columnwidth]{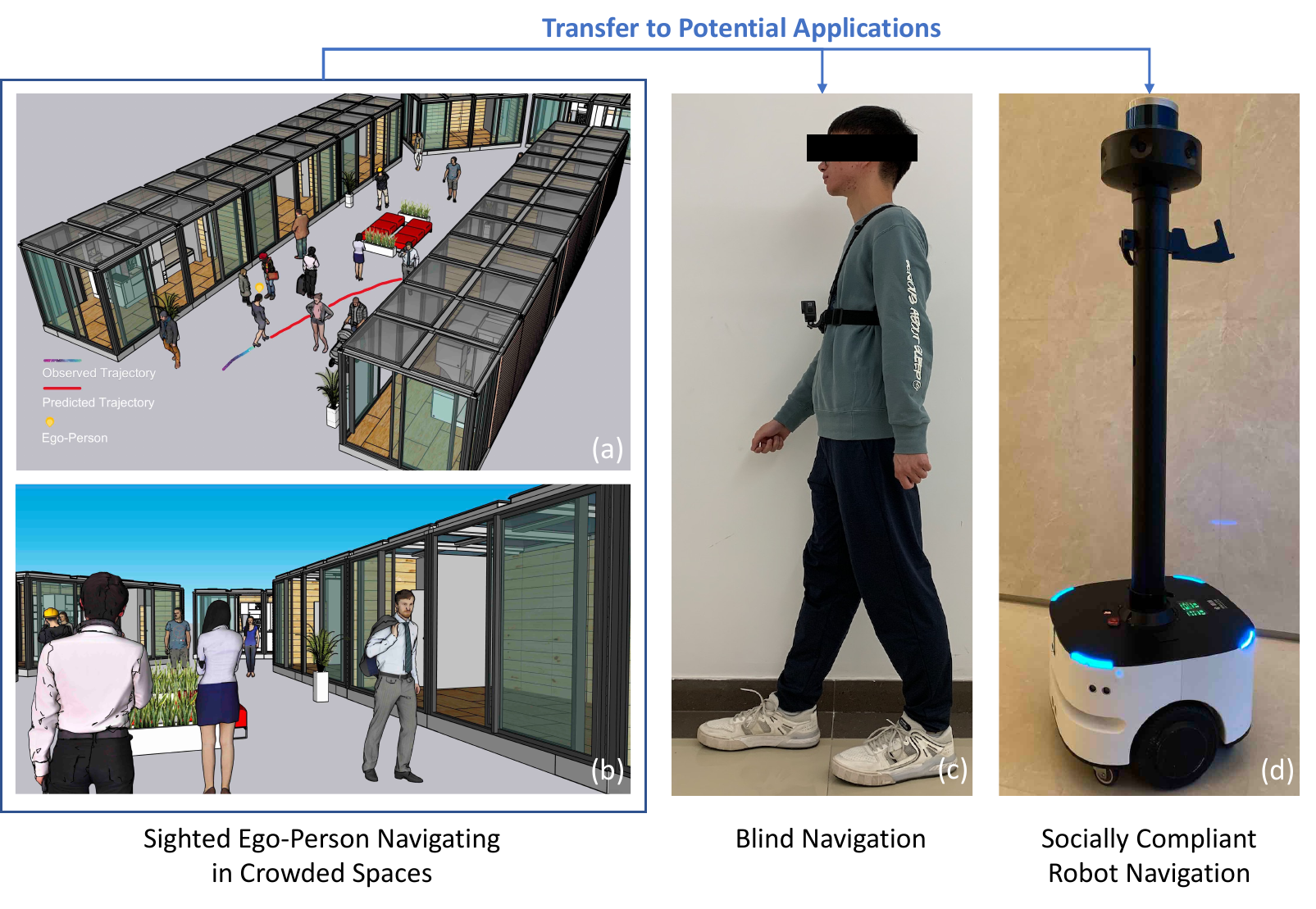}}
\caption{Illustration of the egocentric human trajectory forecasting. (a) Global view of the ego-person's trajectory: an ego-person (a person with a wearable camera) is walking around in a crowded space (e.g., in a shopping mall). The wearable camera, which has an egocentric view as shown in (b), continuously captures close-up details of the wearer's surroundings. In this work, we aim to predict the ego-person's trajectory by only using the contextual cues captured by the wearable camera and the ego-person's past trajectory. The learned egocentric trajectory forecasting ability can be transferred to downstream applications such as assisting blind navigation (c) and teaching robots to navigate in a socially compliant way (d).}
\label{fig:illustration}
\end{figure}

\IEEEPARstart{E}{gocentric} perception has a pivotal role in agent navigation, whether it is a robot, a human, or an autonomous vehicle.  With on-board cameras, forecasting the trajectory of an ego-vehicle, a mobile robot, or their surrounding agents has been actively studied in the fields of autonomous driving and robot navigation to enable better motion planning and to reduce the chances of collision~\cite{everett2018motion,chen2019crowd,marchetti2020multiple}. This provides a new insight for blind navigation~\cite{wang2017enabling,ohnbar2018personalized,zhao2019designing}. With a wearable camera being the proxy of human eyes and an algorithm (learned from sighted people walking with a wearable camera) forecasting the optimal trajectory for navigation in the next few seconds, a blind or visually impaired person can navigate through a space independently in a safer and more natural way (Fig.~\ref{fig:illustration}c).
With a set of synchronized sensors, the predicted optimal trajectory that the user shall follow can be communicated to him/her via vibration~\cite{wang2017enabling} or audio~\cite{google2021}. On the other hand, humans tend to unconsciously follow certain social rules, such as maintaining a comfortable distance with others, and keeping following the people in front of them if not in a rush while walking in crowded spaces. These socially polite ways of navigation are common in human navigation. For a mobile robot, if it can imitate such human navigation behaviours, the robot could better blend itself into human-centric environments while moving in shared spaces with humans (Fig.~\ref{fig:illustration}d). The human navigation behaviors distilled from the egocentric perception data of humans navigating in the real world therefore has the potential to be transferred to mobile robots, enabling socially compliant robot navigation~\cite{kretzschmar2016socially}. In the context of navigation, egocentric trajectory forecasting is similar to motion planning. However, forecasting a camera wearer's trajectory can also find itself useful in other contexts such as AR/VR, by providing a user with a better interactive and immersive experience.

However, such data is currently lacking and limited research has been carried out in forecasting the trajectory of the camera wearer~\cite{park2016egocentric,singh2016krishnacam,bertasius2018egocentric,su2017predicting}. Most human trajectory forecasting methods are targeted for scenarios captured from a bird-eye view or by a static camera~\cite{alahi2016social,gupta2018social,sadeghian2019sophie,kosaraju2019social,liang2019peeking,liang2020garden,mangalam2020not,shafiee2021introvert,yu2020spatio,yuan2021agentformer}. Different from these views, egocentric view brings up the opportunity of understanding how humans perceive the surroundings, and initiate/execute the responsive actions. Such embodied knowledge fully considers the relationship between the agent itself and the close-up contextual information, such as the poses of nearby people and their trajectories, and the sizes of obstacles. These close-up details are crucial for forecasting a safe trajectory. Nevertheless, to the best of our knowledge, there is currently no effort that has been made to forecast the camera wearer's trajectory with these close-up contextual cues. From the perspective of applicability, it is also more practical to implement visual navigation assistance using an egocentric view than using a bird-eye or static camera view.

\begin{figure}[!t]
\centerline{\includegraphics[width=\columnwidth]{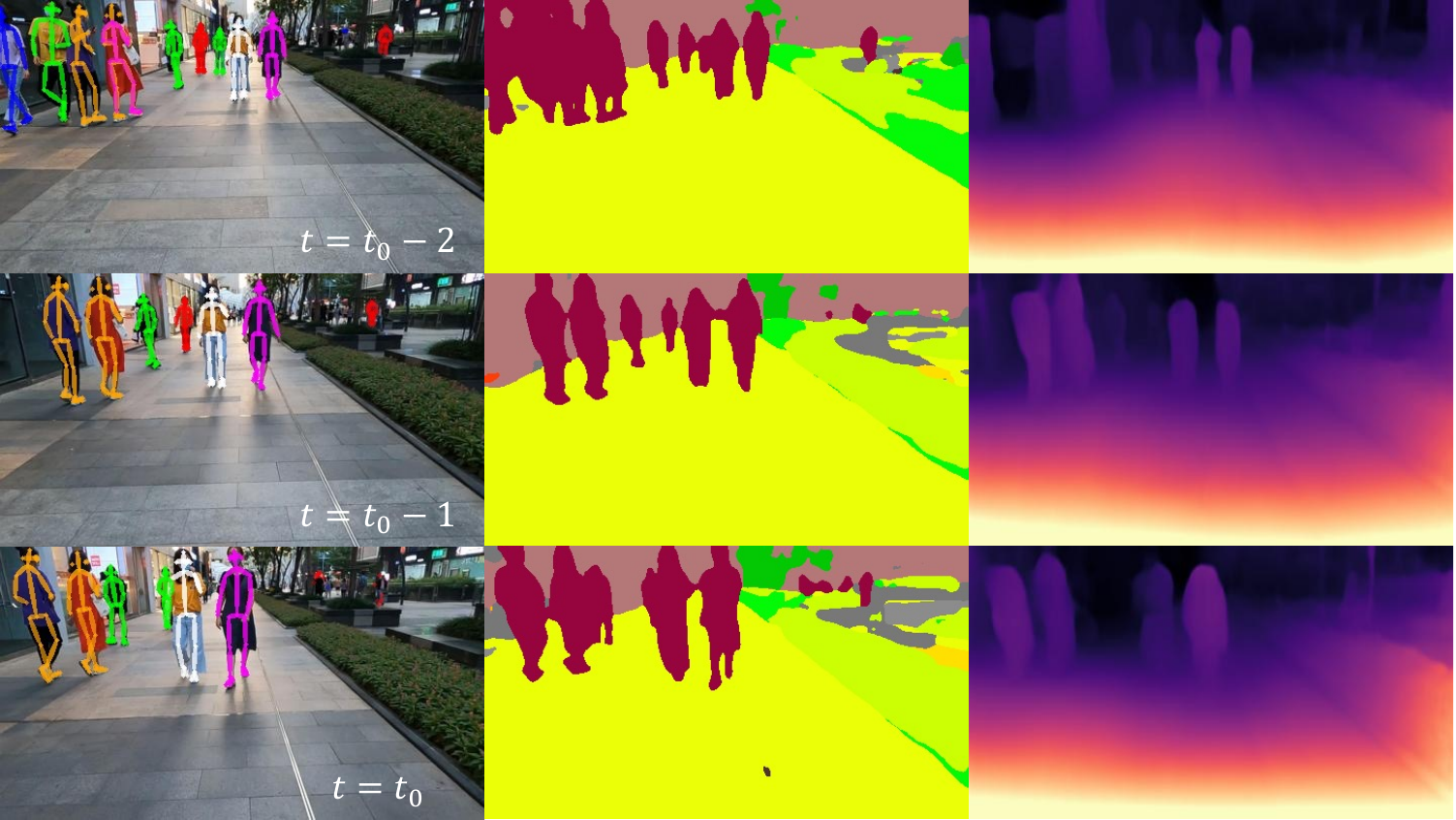}}
\caption{Visual cues for forecasting the trajectory of the camera wearer. (Left) the past trajectories of nearby people (formed by using their poses at each time point. We have also investigated the use of their center body points or bounding boxes to form their trajectories in the 2D pixel space). (Middle) the segmented semantics of the scene at each time point in the observation period. (Right) the estimated depth of the scene at each time point in the observation period.}
\label{fig:data_sample}
\end{figure}

In light of these, in this work, we propose and address the task of forecasting the trajectory of an egocentric camera wearer in crowded spaces. Fig.~\ref{fig:illustration}a and Fig.~\ref{fig:illustration}b provide an illustration of this task. Specifically, given a period of observation, the task is to forecast the camera wearer's trajectory (expressed in the 3D world coordinates with both position and orientation) using cues available in the observation period. We adopt the past trajectory of the camera wearer as the basic input, and meanwhile, utilize the contextual information (as shown in Fig.~\ref{fig:data_sample}), such as the trajectories of nearby people, and the scene semantics or the depth of the scene, to forecast the trajectory of the camera wearer. To this end, an encoder-decoder structure is proposed, in which multiple encoder streams are used to encode different input modalities, and a novel cascaded cross-attention mechanism is then used to fuse the resulting encodings. The decoder receives the fused encodings, and then decodes the future trajectory in an autoregressive manner. The effectiveness of using different modalities has been thoroughly studied in this work. We also implemented socially compliant navigation on a real robot to demonstrate the generalization capability of our model.

In a nutshell, the contributions of this work include:

\begin{itemize}
    \item A novel in-the-wild egocentric human trajectory forecasting dataset has been constructed. The dataset can be downloaded from~\url{https://github.com/Jianing-Qiu/TISS}.
    \item Four different types of modalities are proposed and their effectiveness has been evaluated for egocentric human trajectory forecasting. 
    \item A Transformer-based encoder-decoder framework integrated with a novel cascaded cross-attention mechanism has been designed to forecast the trajectory of the camera wearer in the egocentric setting.
\end{itemize}

\section{Related Work}

In this section, we review prior work in human trajectory prediction, including in both non-egocentric and egocentric scenarios. 

\subsection{\textbf{Non-Egocentric} Human Trajectory Prediction}

Most human trajectory prediction studies are focused on scenes captured from a bird-eye view or by a static camera, i.e., non-egocentric scenarios, with targeted applications, such as intelligent surveillance systems and crowd behavior monitoring. The majority of the work considers modelling pedestrian walking dynamics to approach this task. Early work resorted to hand-craft features such as social force~\cite{helbing1995social}. Recent studies have shifted to neural network-based approaches. Typically, these approaches consider modeling the social interactions between pedestrians to predict a pedestrian's trajectory~\cite{alahi2016social,gupta2018social,sadeghian2019sophie,kosaraju2019social}, or jointly modeling future trajectory with future activities~\cite{liang2019peeking} or with destinations~\cite{mangalam2020not}. Liang et al.~\cite{liang2020garden} used multi-scale location decoders to predict multiple future trajectories. Shafiee et al.~\cite{shafiee2021introvert} proposed a conditional 3D attention mechanism for pedestrian trajectory prediction.

Transformer~\cite{vaswani2017attention} recently has established new state-of-the-arts in a series of vision and language tasks. Few studies have also exploited the transformer-based architecture for pedestrian trajectory prediction. STAR~\cite{yu2020spatio} predicts pedestrian trajectories with only the attention mechanism, which is achieved by a graph-based spatial transformer and a temporal transformer. AgentFormer~\cite{yuan2021agentformer}, a transformer-based framework, jointly models temporal and social dimensions in human motion dynamics to predict future trajectories. Our model is also based on transformer, but we differ from them in that 1) we target for egocentric scenarios, and 2) our model encodes multiple modalities with a novel cascaded cross-attention mechanism, whereas their models are designed to use the past trajectories as the only cue for the future trajectory prediction.

\subsection{\textbf{Egocentric} Human Trajectory Prediction}

In egocentric scenarios, based on the types of used visual sensors, human trajectory prediction can be categorized into the vehicle-mounted camera scenario and the wearable camera scenario.

\subsubsection{Vehicle-Mounted Camera Scenario} In the field of autonomous driving, pedestrian trajectory prediction has been actively researched~\cite{kooij2014context,bhattacharyya2018long,styles2019forecasting,chandra2019traphic,makansi2020multimodal,mangalam2020disentangling}. Some research has also studied the prediction of a pedestrian's intent of crossing or not crossing the road~\cite{gujjar2019classifying,liu2020spatiotemporal}, and the detection of traffic accident by predicting pedestrians' future locations~\cite{yao2019unsupervised}. As the vehicles are mainly on the road, in the images/videos captured by a vehicle-mounted camera, the pedestrians often appear on the sidewalks, and therefore are mainly in the left or right section of the images/videos, or in the middle when the vehicle is at an intersection. These scenes hence differ from those captured by a wearable camera, as the places the camera wearer can visit and enter are different from vehicles and the locations of nearby people can be in close proximity to the camera wearer. For example, the camera wearer can walk in a long distance following the person in front, which can rarely be found in vehicle-mounted camera scenarios.

\subsubsection{Wearable Camera Scenario} Among all the works using wearable cameras, Yagi et al.~\cite{Yagi_2018_CVPR} first proposed the task of future person localization in egocentric videos. In their work, a multi-stream 1D  convolution-deconvolution (1D Conv-DeConv) network was designed to utilize a pedestrian's past poses, locations and scales, as well as ego-motion to infer his/her future locations. In~\cite{styles2020multiple}, a GRU-CNN-based framework was proposed to conduct future person localization. An LSTM-based encoder-decoder framework was utilized in~\cite{qiu2021indoor} to predict human trajectory in indoor environments. The above work, similar to those in the vehicle-mounted camera scenario, is focused on predicting the trajectories of the pedestrians captured by egocentric cameras, which is different from our work. Our work is to predict the trajectory of the camera wearer rather than those of pedestrians, which could benefit the blind navigation as well as the socially compliant robot navigation.

Nevertheless, little research has been carried out in predicting the camera wearer's trajectory. An early attempt was made by Park et al.~\cite{park2016egocentric}, who used stereo images and an EgoRetinal map to predict the camera wearer's future path. Given a single egocentric image, a nearest-neighbor based search approach was used in~\cite{singh2016krishnacam} to predict the future trajectory. The search was based on the measured deep-feature similarity between images. Camera wearer trajectory prediction has also been studied in the context of playing basketball. Bertasius et al.~\cite{bertasius2018egocentric} proposed to use a future CNN and a goal verifier network followed by an inverse synthesis procedure to estimate a basketball player's motion. In~\cite{su2017predicting}, egocentric videos were used and basketball players' trajectories were predicted by a retrieval-based method. Although our work also predicts the trajectory of the camera wearer, we differ from them in that 1) we model the interactions between the camera wearer and the people in his/her close vicinity in both crowded indoor and outdoor spaces; 2) rich visual features, such as scene semantics and depth information, are leveraged to aid trajectory prediction; 3) and without bells and whistles, our method, based on a simple encoder-decoder structure and a novel cascaded cross-attention mechanism, has shown to be effective in fusing multiple contextual cues and accurately predicting the trajectory of the camera wearer.

\section{Method}
\label{sec:method}

\subsection{Problem Formulation}

We formulate the task of predicting the trajectory of an egocentric camera wearer as to learn a function $\phi_\theta: (\mathcal{Y}, \mathcal{Z}) \to \hat{\mathcal{Y}}$, which takes the past trajectory $\mathcal{Y}$ of the camera wearer and some contextual cues $\mathcal{Z}$ as inputs, and outputs the camera wearer's future trajectory $\hat{\mathcal{Y}}$.

\begin{figure}
     \centering
     \begin{subfigure}[b]{0.48\columnwidth}
         \centering
         \includegraphics[width=\columnwidth]{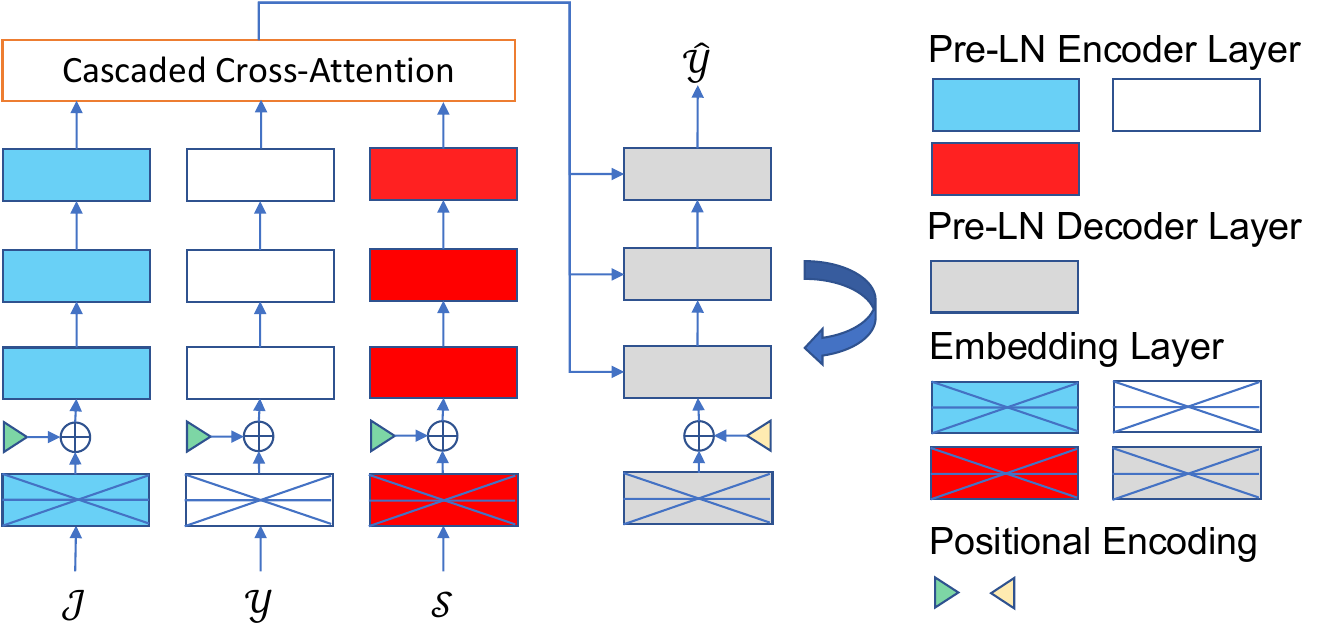}
         \caption{Transformer-based encoder-decoder structure}
         \label{fig:encoder_decoder}
     \end{subfigure}
     \hfill
     \begin{subfigure}[b]{0.48\columnwidth}
         \centering
         \includegraphics[width=\columnwidth]{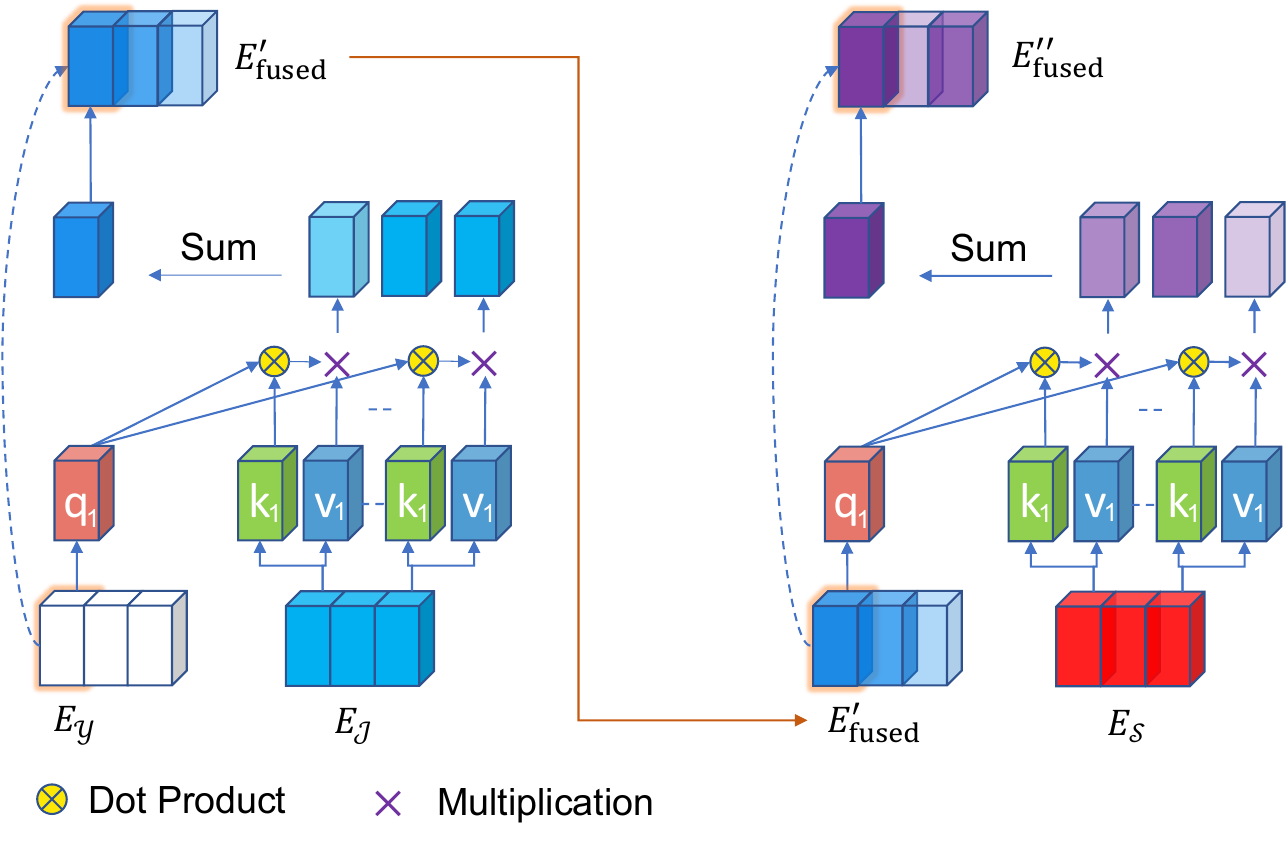}
         \caption{Cascaded cross-attention}
         \label{fig:cxa}
     \end{subfigure}
        \caption{(a) Overview of the model structure. The model is composed of three encoder streams, each with three Pre-LN Transformer layers to encode an input modality, and a single decoder stream to decode the future trajectory of the camera wearer in an autoregressive manner. A cascaded cross-attention mechanism is designed to fuse the encodings of multiple modalities. (b) Illustration of the cascaded cross-attention mechanism.}
        \label{fig:network}
\end{figure}

\subsection{Ego-Trajectory Forecasting with Multiple Modalities}\label{subsec: ego-traj_forecasting_with_multi-modalities}
People navigate in crowded spaces with social etiquette, and the movement of each individual is influenced by those of the nearby people. Therefore, in predicting the trajectory of a camera wearer, the trajectories of people in his/her close vicinity can play an important role. Hence, we use the nearby people's trajectories $\mathcal{J}$ as one of the cues for forecasting the camera wearer's trajectory. In addition, understanding the scene semantics (e.g., walkable areas such as the floor, and obstacles such as walls) can also aid trajectory forecasting. We therefore use scene semantic segmentation $\mathcal{S}$ as another cue for camera wearer trajectory forecasting. Hence, we have $\mathcal{Z} = [\mathcal{J, S}]$.

We have also experimented on the depth information, and used depth estimation $\mathcal{D}$ as an alternative to the scene semantic segmentation, i.e., $\mathcal{Z} = [\mathcal{J, D}]$. As the methodologies of processing and fusing scene segmentation and depth estimation with other two modalities are similar, in this section, we mainly focus on the use of the past trajectory $\mathcal{Y}$ of the camera wearer, those of nearby people $\mathcal{J}$, and the scene semantic segmentation $\mathcal{S}$. Experimental results of using the depth estimation will be present in Section~\ref{subsubsec:abl_studies}.

Concretely, given a period of observation (past) $\text{T}_{\text{obs}}$, the camera wearer's past trajectory $\mathcal{Y} \in \mathbb{R}^{\text{T}_{\text{obs}} \times 7}$ is formed chronologically using the camera position ($\bm{p} \in \mathbb{R}^{\text{T}_{\text{obs}} \times 3}$) and orientation ($\bm{o} \in \mathbb{R}^{\text{T}_{\text{obs}} \times 4}$, as quaternion) at each time point $t \in (t_{0} - \text{T}_{\text{obs}} + 1, ..., t_{0})$ where $t_{0}$ is the end of the observation period. The past trajectories of nearby people $\mathcal{J} = \sum_{n=1}^{N} J_{n}$ where $N$ is the number of nearby people and $J_{n} \in \mathbb{R}^{\text{T}_{\text{obs}} \times d}$ is each nearby person's trajectory projected in the 2D pixel space (the viewpoint of the camera wearer) during the time period $t \in (t_{0} - \text{T}_{\text{obs}} + 1, ..., t_{0})$, and $d$ is the dimension of the cue we use to represent the nearby people's trajectories. We mainly use a nearby person's pose to form his/her trajectory, but the use of his/her center body point, as well as his/her bounding box has also been investigated. For scene semantic segmentation, we encode the corresponding segmentation mask at each time point in the observation period into a $k$-dimensional vector, i.e., $\mathcal{S} \in \mathbb{R}^{\text{T}_{\text{obs}} \times k}$. Similarly, if the depth images are used, we have $\mathcal{D} \in \mathbb{R}^{\text{T}_{\text{obs}} \times k}$.

Finally, given a prediction period $\text{T}_{\text{pred}}$, the future trajectory of the camera wearer $\mathcal{\hat{Y}} \in \mathbb{R}^{\text{T}_{\text{pred}} \times 7}$ can be predicted using $\phi_\theta (\mathcal{Y}, \mathcal{Z})$ where $\mathcal{Z} = [\mathcal{J, S}]$ or $\mathcal{Z} = [\mathcal{J, D}]$.

\begin{algorithm}[t] 
\SetAlgoLined
\SetKwData{Wq}{$W_q$}
\SetKwData{Wk}{$W_k$}
\SetKwData{Wv}{$W_v$}
\SetKwData{Ey}{$E_\mathcal{Y}$}
\SetKwData{Ej}{$E_\mathcal{J}$}
\SetKwData{Es}{$E_\mathcal{S}$}
\SetKwData{Efused}{$E_\text{fused}$}
\SetKwData{Eothers}{$E_\text{others}$}
\SetKwData{Efusedprime}{$\Efused^{\prime}$}
\SetKwData{E}{$E$}

\SetKwFunction{LN}{LN}
\SetKwFunction{Softmax}{softmax}

\KwData{\Ey, \Ej, and \Es}
\KwResult{\Efused: the fused encodings of all input modalities.}
\tcc{\LN is Layer Normalization. \Wq, \Wk, and \Wv are learnable matrices. $1/\sqrt{d_{k}}$ is the scaling factor.}
$\Efused \leftarrow \Ey$\;
$\Eothers \leftarrow [\Ej, \Es]$\;
\For{\E in \Eothers}{

    $\Efusedprime \leftarrow$ \LN{$\Efused$}\;
    $\Efused \leftarrow$ \Softmax{$\Wq\Efusedprime(\Wk\E)^{T}/\sqrt{d_{k}}$}$\Wv\E$ $+ \Efused$\;

}
 \caption{Cascaded Cross-Attention Mechanism}
 \label{algo:cascaded_cross_attention}
 
\end{algorithm}

\subsection{Model Structure}

We implement $\phi_\theta$ as a Transformer based encoder-decoder neural network model containing a novel cascaded cross-attention mechanism to fuse encodings of different modalities. Fig.~\ref{fig:network} shows the model structure. Each modality is encoded by a separate encoder stream, and each stream contains three Pre-LN Transformer layers~\cite{xiong2020layer}. After encoding, the inputs $\mathcal{Y, J}$ and  $\mathcal{S}$ are transformed into $E_\mathcal{Y}$, $E_\mathcal{J}$, and $E_\mathcal{S}$, respectively. We then utilize a cascaded cross-attention mechanism (see Algorithm~\ref{algo:cascaded_cross_attention}) to fuse them. The fused encoding $E_\text{fused}$ is then fed into the decoder. In inference, we use greedy decoding to decode the future trajectory, and based on the experiments, using greedy decoding in training is able to increase the trajectory prediction accuracy during inference. This increase is also observed in~\cite{yuan2021agentformer}. Therefore, in both training and inference, the decoder decodes the future trajectory of the camera wearer in a greedy autoregressive manner. $\mathcal{L}_{2}$ loss is used during model training. We name our model as CXA-Transformer (Cascaded Cross-Attention Transformer).

\section{Experiment}
\label{sec:experiment}

\subsection{Dataset}

A novel egocentric human trajectory forecasting dataset has been constructed. To the best of our knowledge, this new dataset is the first of its kind in the community.

\subsubsection{Data Collection}
Camera wearers were wearing a GoPro Hero 9 Black camera while walking around in crowded spaces, such as a large shopping mall. Six unique areas were chosen to collect the data. Both indoor and outdoor scenarios were covered. For outdoor scenarios, data was collected in busy city areas where vehicles are not allowed to enter. The camera was mounted on the chest of the wearer and facing forward. The camera recorded egocentric videos with a resolution of $1920 \times 1080$ at 30 fps.

\begin{figure}
     \centering
     \begin{subfigure}[b]{0.50\columnwidth}
         \centering
         \includegraphics[width=\columnwidth]{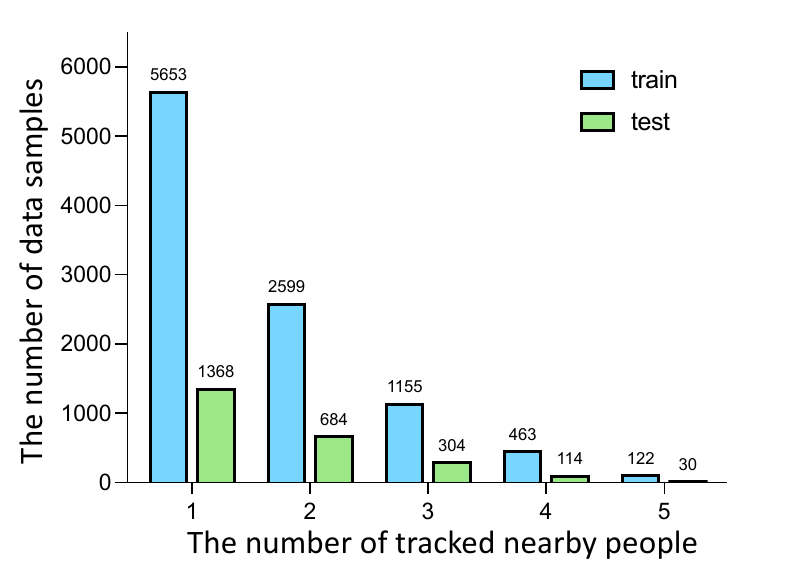}
         \caption{}
         \label{fig:num_sp}
     \end{subfigure}
     \hfill
     \begin{subfigure}[b]{0.48\columnwidth}
         \centering
         \includegraphics[width=\columnwidth]{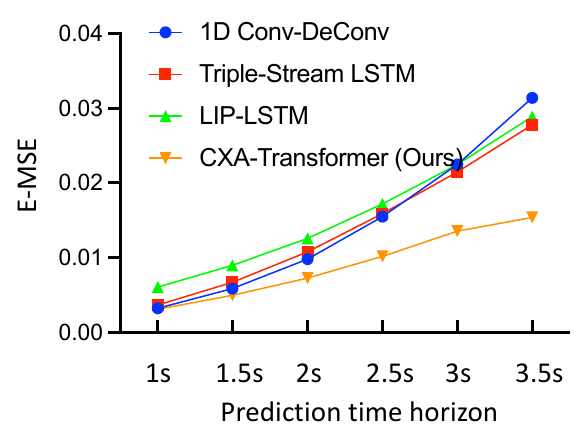}
         \caption{}
         \label{fig:diff_horizon}
     \end{subfigure}
        \caption{(a) The number of data samples with respect to the different number of tracked nearby people. (b) Comparison between different methods on predicting the trajectory at different time horizons.}
        \label{fig:num_sp_diff_horizon}
\end{figure}

\begin{figure}
     \centering
     \begin{subfigure}[b]{0.485\columnwidth}
         \centering
         \includegraphics[width=\columnwidth]{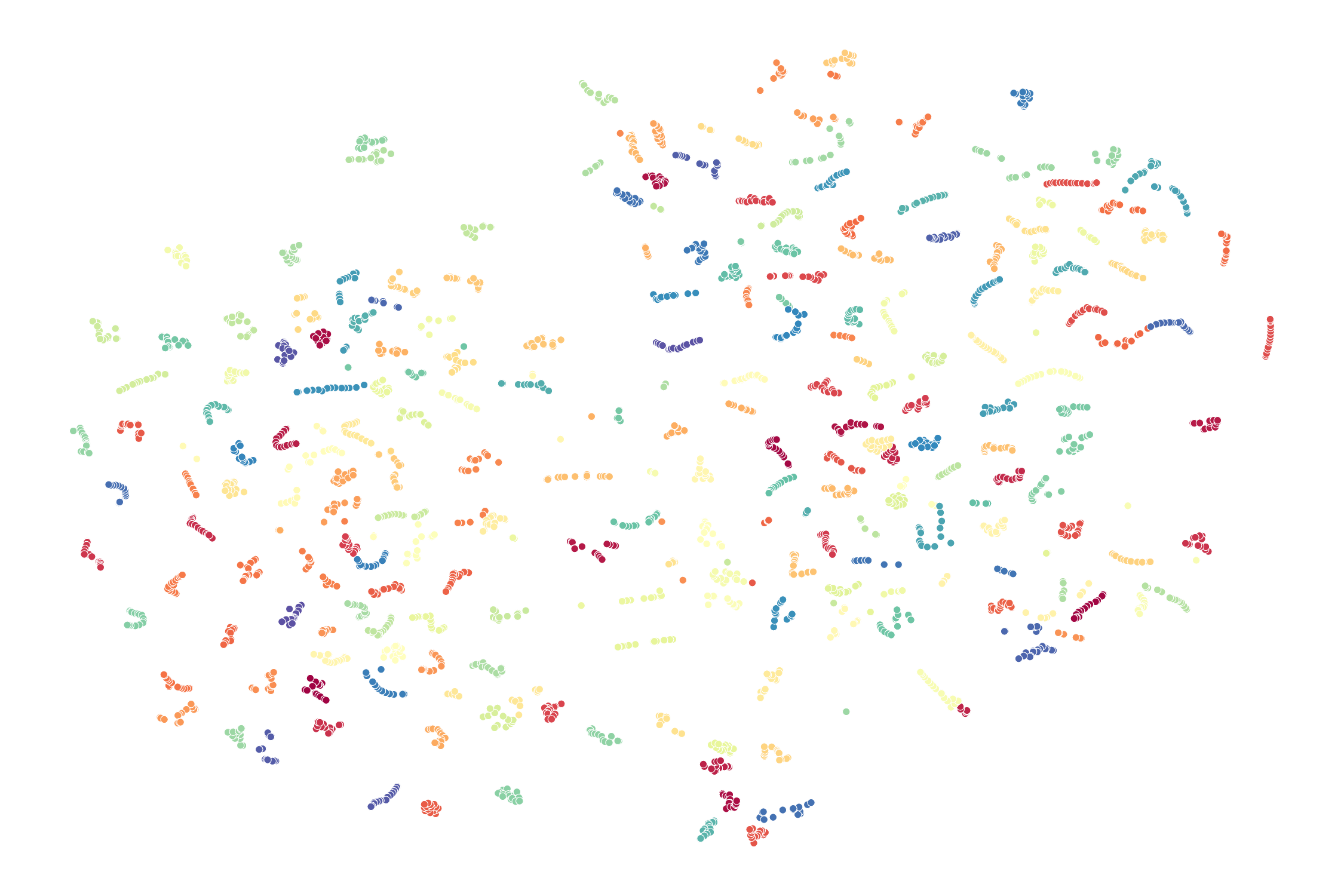}
         \caption{}
         \label{fig:ss_tsne}
     \end{subfigure}
     \hfill
     \begin{subfigure}[b]{0.485\columnwidth}
         \centering
         \includegraphics[width=\columnwidth]{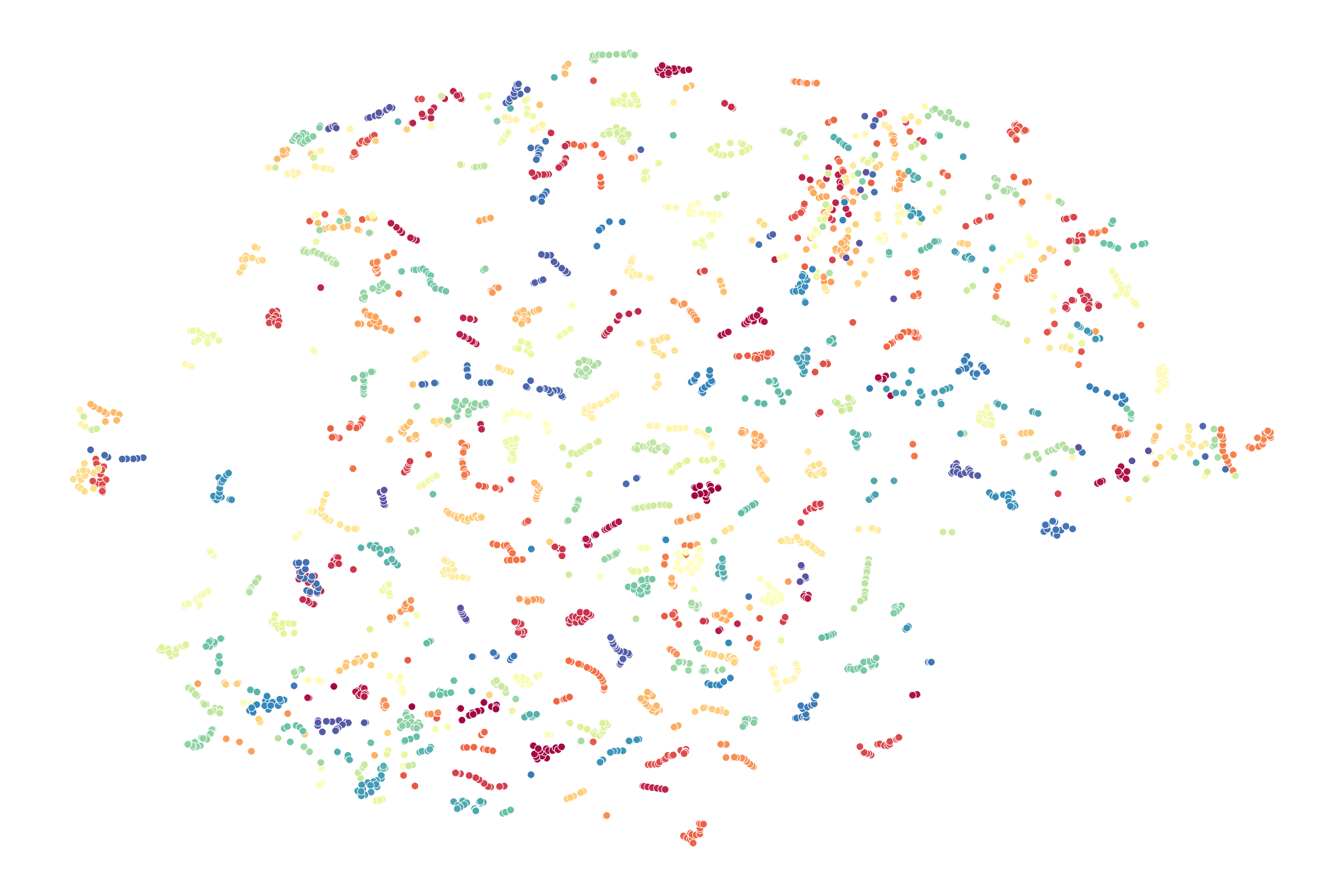}
         \caption{}
         \label{fig:depth_tsne}
     \end{subfigure}
        \caption{t-SNE visualization of the encodings of scene semantic segmentation masks (a) and depth images (b). Each cluster is the encodings of the segmentation masks or the depth images of one trajectory in the observation period (best viewed in zoom in). Each encoding is represented by a dot. Same color means they are from the same trajectory. For visualization purpose, 1) we use 15 frames for each trajectory, i.e, 10fps, and the used frames of 2fps are within these 15 frames; 2) 200 samples are randomly chosen from the test set, as shown in (a) and (b). It can be observed that the encodings learned from the designed autoencoder are discriminative between different trajectories, and the encodings of scene semantic segmentation are more discriminative than those of depth images, which is also confirmed by the experiments discussed in Section~\ref{subsubsec:abl_studies}}
        \label{fig:tsne}
\end{figure}

\begin{figure*}[!t]
\centerline{\includegraphics[width=\textwidth]{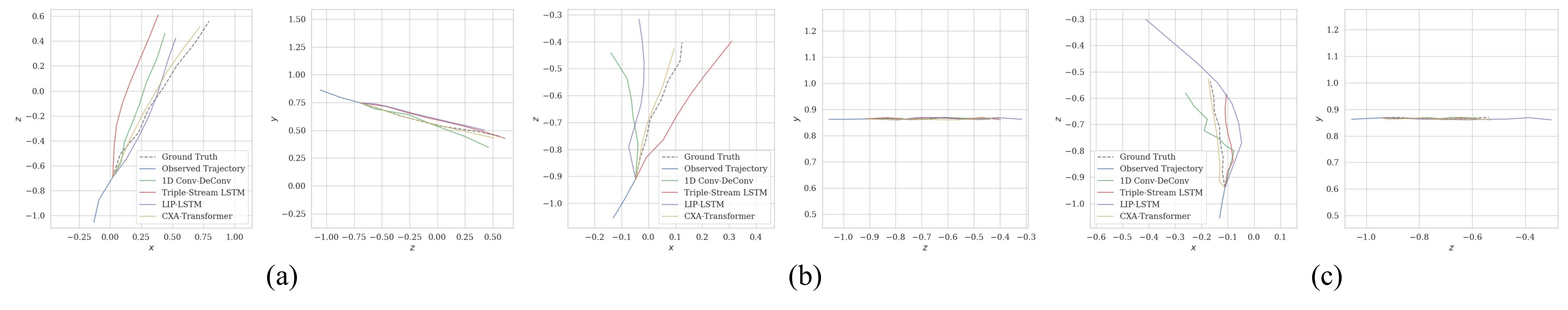}}
\caption{Visualization of the predicted trajectories of different methods, the ground truth, and the observed trajectory. The predicted trajectory of our CXA-Transformer better aligns with the ground truth compared to those of baselines. We show two views for each of the three examples. Left: x-z horizontal plane. Right: z-y vertical plane.}
\label{fig:vis_trajs}
\end{figure*}

\subsubsection{Data Pre-Processing}

As our data was collected in the real world, including both indoor and outdoor environments, it is not feasible to use GPS (considering it is not available in the indoor environment) and also not feasible to use motion capture systems such as Vicon, which are installed in a studio or a laboratory, to obtain the camera wearer's trajectories. Using SLAM becomes the option to obtain the camera wearer's trajectories under our data collection setting. ORB-SLAM3~\cite{campos2021orb} is one of cutting-edge SLAM algorithms that can precisely output an camera's locations in the 3D world. Its average absolute trajectory error is 0.041m tested on the EuRoC dataset~\cite{burri2016euroc}. We therefore ran ORB-SLAM3~\cite{campos2021orb} to obtain the ground truth trajectories of the camera wearer from the recorded egocentric monocular RGB videos. Note that the raw ground truth trajectories are only measured with reference to the map initialized by the ORB-SLAM3 algorithm. Their length is not in the real-world scale, but their shape is the same as in the real world. We then used ffmpeg to extract frames from the videos and set the resolution of each frame to $480 \times 270$. AlphaPose~\cite{xiu2018poseflow} was then used to detect and track the nearby people appearing in each frame. For each tracked nearby person, we used his/her detected pose in each frame to form his/her trajectory in the 2D pixel space. We also used the center of his/her neck and hip keypoints to represent his/her center body point in each frame. The center body point and the detected bounding box of a nearby person are used as alternatives to the pose to form his/her trajectory. PSPNet~\cite{zhao2017pyramid} was used to segment scene semantics. The parsed scene semantics of each frame is represented by a segmentation mask with the same size as the RGB frame. After segmentation masks were obtained, we trained an autoencoder to encode each mask into a $k$-dimensional vector. Monodepth2~\cite{monodepth2} was used to estimate the depth from the monocular RGB frames. Similar to the segmentation mask, we trained a separate autoencoder to encode each depth image into a $k$-dimensional vector.

The timestamps of the camera wearer's trajectory are aligned with those of video frames, and we reduce the frame rate to 2fps. Each trajectory of the camera wearer is 5 seconds long (i.e., 10 time points), and records a sequence of camera poses expressed in relative to the initial pose of that sequence. 

In total, the dataset contains 12,492 unique trajectories, and we used 9,992 for training and the rest 2,500 for testing. We limited the maximum number of tracked nearby people to 5 in the dataset. Note that this is not the actual number of people in the environment. Fig.~\ref{fig:num_sp} shows the number of data samples with respect to the different number of tracked nearby people. We use t-SNE~\cite{van2008visualizing} to visualize the encoded vectors (by the respective autoencoder) of segmentation masks and depth images in Fig~\ref{fig:tsne}. We name our dataset as TISS (the initial of four institutions with which the authors are affiliated).

\subsection{Implementation Details}

Our model was implemented using PyTorch. We used 4 attention heads in the Pre-LN Transformer layers and the cascaded cross-attention module. Hidden layer dimension of the Pre-LN Transformer's feed forward sub-layers was set to 2,048. The embedding layers transform modalities of different dimensions to the same dimension of 512.  The implementation of PSPNet and its pre-trained model were adopted from \cite{zhou2018semantic}. We set the observation time $\text{T}_{\text{obs}}$ to 3 time points (at 2fps, it is equal to 1.5 s) and the prediction time $\text{T}_{\text{pred}}$ to 7 time points (at 2fps, it is equal to 3.5 s). 
We used 26 body keypoints to represent a nearby person's pose. We used zero padding to pad those data samples which have the number of tracked nearby people less than 5. Therefore, $d$ was set to 260 when body pose was used, and $d$ was 10 and 20 respectively, if the center body point and the bounding box were used. $k$ was set to 648. Each modality was normalized in order to accelerate the convergence of the model. We trained our model for 300 epochs using Adam optimization~\cite{kingma2014adam}. The batch size was set to 1,024 with a learning rate of 5e-5.

\subsection{Baselines}

We compared the performance of egocentric human trajectory prediction of our \textbf{CXA-Transformer} model with the following baseline methods, all of which are able to take multiple modalities simultaneously as their inputs:

\begin{itemize}
    \item \textbf{1D Conv-DeConv}~\cite{Yagi_2018_CVPR}: A multi-stream 1D convolution-deconvolution framework originally proposed for future pedestrian localization in egocentric videos. We adapt it to be compatible with different modalities and their different dimensions in our dataset.
    \item \textbf{Triple-Stream LSTM}: An LSTM-based encoder-decoder framework with three encoder streams and one decoder stream. Different input modalities are merged using two fully connected layers (one for hidden states, and the other for cell) before being fed into the decoder.
    \item \textbf{LIP-LSTM}~\cite{qiu2021indoor}: A single stream LSTM-based encoder-decoder framework originally proposed for predicting the location and movement trajectory of pedestrians captured in egocentric videos. We concatenate input modalities and then feed the resulting vector to its encoder. Future trajectory is then predicted by its decoder. 
\end{itemize}

The above neural network-based baselines were also trained with $\mathcal{L}_{2}$ loss. Apart from them, we also implemented two simple trajectory predictors that take only the past trajectory as the input to predict the future trajectory. The first one is based on random sampling. We randomly sample the position difference and orientation difference between the time point $t+1$ and the time point $t$ in the future based on the differences measured between the time points in the observation period $\text{T}_{\text{obs}}$. The minimum and maximum values of those differences in the past are used as the minimum and maximum range for the sampling. The future trajectory is then obtained by iteratively adding the randomly sampled position and orientation differences to the end of the observed trajectory. The second one is a Gaussian Process Regression-based method (GPR), which was trained using the trajectories in the training set.

\begin{table}[]
\centering
\caption{Comparison with the State-of-the-Art Methods}
\label{tab:comp_sota}
\begin{tabular}{@{}lcccc@{}}
\toprule
Method             & P-MSE         & O-MSE      & E-MSE   & F-MSE       \\ \midrule
1D Conv-DeConv~\cite{Yagi_2018_CVPR}     & 7.09e-2          & 1.77e-3          & 3.14e-2       & 8.46e-2   \\
Triple-Stream LSTM & 6.40e-2          & 7.37e-4          & 2.78e-2    & 6.59e-2      \\
LIP-LSTM~\cite{qiu2021indoor}           & 6.67e-2          & 4.54e-4          & 2.89e-2       & 6.69e-2   \\ \midrule
CXA-Transformer (Ours)    & \textbf{3.54e-2} & \textbf{4.27e-4} & \textbf{1.54e-2} & \textbf{4.20e-2} \\ \bottomrule
\end{tabular}%
\end{table}

\subsection{Evaluation Metrics}

We used MSE to measure the future trajectory prediction error of each method. We first measure the MSE error over the entire predicted trajectory, denoted as E-MSE (i.e., 7 time points in the case of predicting 3.5s into the future). We also separately report the MSE error of position and orientation prediction, denoted as P-MSE and O-MSE, respectively. In addition, we also compared the performance of each method at different prediction horizons, i.e., 1.0s, 1.5s, 2.0s, 2.5s, and 3.0s. Finally, we also report the MSE error of the end point, i.e, the final prediction time point, which is denoted as F-MSE.

\subsection{Results}

We first report the E-MSE error of the two simple trajectory predictors. We executed the random sampling one for 10 times, and its error is 5.93e-2$\pm$2.5e-4. For the GPR model, its E-MSE error is 4.87e-2. We then compare our proposed method with the neural-network based baselines in the following section.

\subsubsection{Comparison with the State-of-the-Art}

Table~\ref{tab:comp_sota} summarizes the results of our method and the state-of-the-art neural network-based baselines on the task of egocentric human trajectory forecasting. Our CXA-Transformer achieves the lowest prediction error in terms of the estimated future position and orientation of the camera wearer. Its E-MSE error is nearly the half of the second best method's error. The F-MSE achieved by our model is also lower than those of baselines. Fig.~\ref{fig:vis_trajs} shows three examples of the predicted trajectories of different methods as well as the ground truth and the observed trajectory. It can be seen that the predicted trajectory of our model aligns better with the ground truth than those of baselines.

\begin{table}[]
\centering
\caption{Comparison between the Proposed Cascaded Cross-Attention (CXA) and Other Fusion Mechanisms}
\label{tab:fusion_compare}
\begin{tabular}{@{}cccc@{}}
\toprule
Fusion Mechanism & P-MSE         & O-MSE      & E-MSE          \\ \midrule
Average Pooling  & 5.79e-2          & 5.83e-4          & 2.51e-2          \\
Max Pooling      & 4.85e-2          & 5.86e-4          & 2.11e-2          \\
Linear Layer     & 4.61e-2          & 5.12e-4          & 2.00e-2          \\ \midrule
CXA (ours)       & \textbf{3.54e-2} & \textbf{4.27e-4} & \textbf{1.54e-2} \\ \bottomrule
\end{tabular}
\end{table}

\begin{table}[]
\centering
\caption{Ablation Studies of Using Different Types of Modalities as the Input to CXA-Transformer}
\label{tab:abl_diff_modality}
\begin{tabular}{@{}lccc@{}}
\toprule
Modality & P-MSE         & O-MSE      & E-MSE          \\ \midrule
$\mathcal{Y}$        & 1.06e-1          & 7.10e-4          & 4.60e-2          \\ \midrule
$\mathcal{Y+C}$      & 1.01e-1          & 6.56e-4          & 4.36e-2          \\
$\mathcal{Y+B}$      & 9.28e-2          & 6.78e-4          & 4.01e-2          \\
$\mathcal{Y+P}$      & 9.06e-2          & 6.31e-4          & 3.92e-2          \\
$\mathcal{Y+S}$     & 6.23e-2          & 6.00e-4          & 2.70e-2          \\
$\mathcal{Y+D}$      & 7.02e-2          & 6.66e-4          & 3.05e-2          \\ \midrule
$\mathcal{Y+C+S}$    & 3.73e-2          & 4.57e-4          & 1.63e-2          \\
$\mathcal{Y+B+S}$    & 3.61e-2          & 4.38e-4          & 1.57e-2          \\
$\mathcal{Y+P+S}$    & \textbf{3.54e-2} & \textbf{4.27e-4} & \textbf{1.54e-2} \\
$\mathcal{Y+C+D}$    & 4.93e-2          & 5.26e-4          & 2.14e-2          \\
$\mathcal{Y+B+D}$    & 4.76e-2          & 5.36e-4          & 2.07e-2          \\
$\mathcal{Y+P+D}$    & 4.81e-2          & 5.37e-4          & 2.09e-2          \\ \bottomrule
\end{tabular}%
\end{table}

Fig.~\ref{fig:diff_horizon} shows the results of predicting future trajectory at various horizons. The proposed CXA-Transformer shows better performance than the baselines at all different prediction horizons. It is worth noting that at very short prediction horizons, e.g., 1.0s, and 1.5s, the prediction error of the CXA-Transformer is not significantly lower than those of the baselines. However, as the prediction horizon increases, the error of our method grows at a slower rate than those of the baselines.

\begin{figure*}
     \centering
     \begin{subfigure}[b]{0.22\textwidth}
         \centering
         \includegraphics[width=\textwidth]{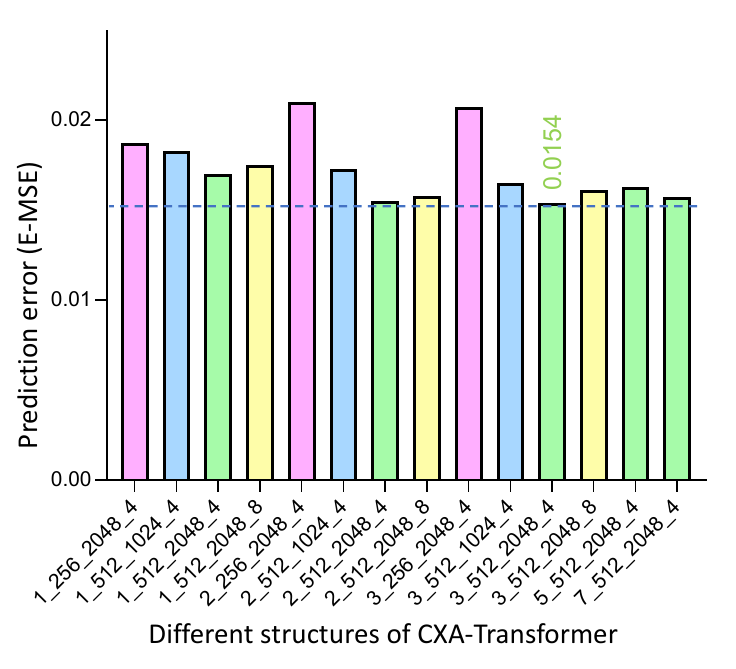}
         \caption{}
         \label{fig:net_structure_params_search}
     \end{subfigure}
     \hfill
     \begin{subfigure}[b]{0.37\textwidth}
         \centering
         \includegraphics[width=\textwidth]{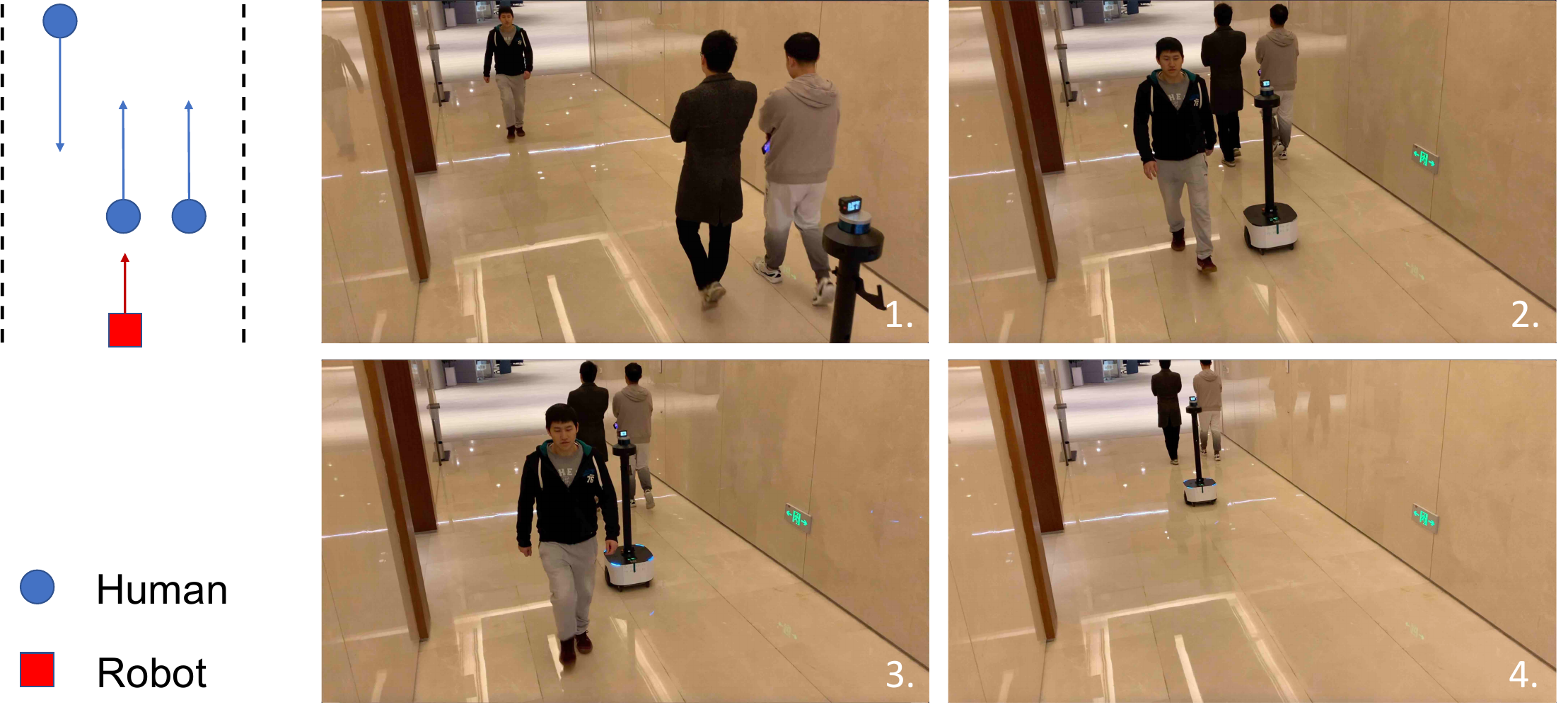}
         \caption{}
         \label{fig:robot_experiment_1}
     \end{subfigure}
     \hfill
     \begin{subfigure}[b]{0.37\textwidth}
         \centering
         \includegraphics[width=\textwidth]{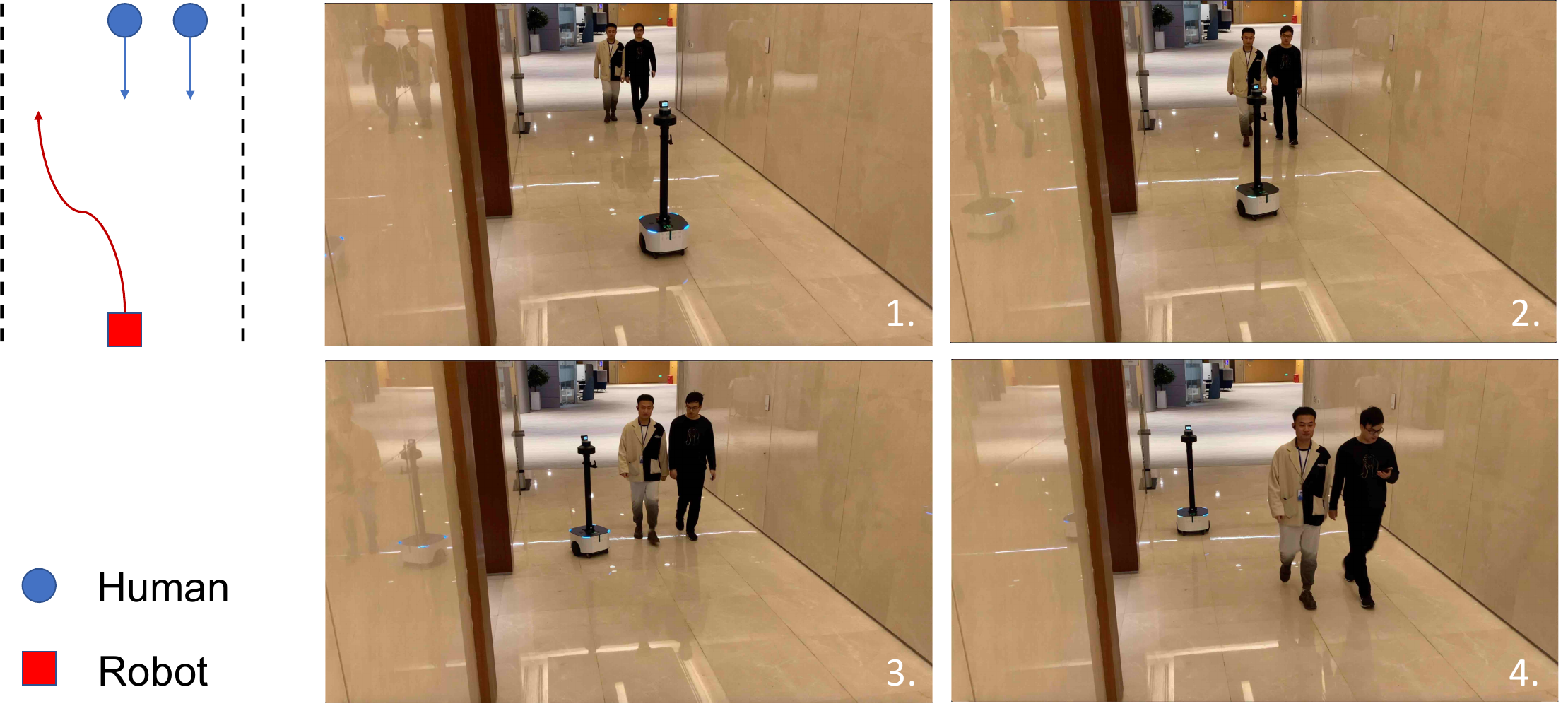}
         \caption{}
         \label{fig:robot_experiment_2}
     \end{subfigure}
     \caption{(a) Accuracy of the CXA-Transformer with the different number of Pre-LN layers (NP), different embedding dimensions (DE), different dimensions of the feed-forward sub-layer (DF), and different number of attention heads (NA). Accuracy is reported with the use of $\mathcal{Y+P+S}$ as the input. X label is represented in the format of NP\_DE\_DF\_NA. (b)-(c) Our model predicts socially compliant trajectories for robot navigation. (b) Robot follows people in front instead of overtaking them. (c) Robot turns left to avoid collision with the people walking towards it.}
     \label{fig:robot_experiment_net_structure_params_search}
\end{figure*}

\subsubsection{Ablation Studies}\label{subsubsec:abl_studies}

\hfill \break \indent
\textbf{Different Fusion Mechanisms:} We replaced the CXA fusion module in our model with average pooling, max pooling, or a linear layer followed by relu activation to fuse the encodings of different input modalities. As shown in Table~\ref{tab:fusion_compare}, our proposed CXA fusion mechanism leads to a better result compared to other traditional fusion mechanisms.

\textbf{Different Modalities:} We further studied the effectiveness of each input modality in predicting egocentric human trajectory. Table~\ref{tab:abl_diff_modality} summarizes the results of using different number of modalities as the input to the CXA-Transformer. Compared to using the camera wearer's past trajectory $\mathcal{Y}$ as the only cue for prediction, adding the trajectories of nearby people (whether using their center body point $\mathcal{C}$, bounding box $\mathcal{B}$, or pose $\mathcal{P}$), or the scene semantics $\mathcal{S}$, or the depth information $\mathcal{D}$ can reduce the prediction error. When adding the cues of both the nearby people and the environment, the prediction error can be further reduced. The CXA-Transformer achieved the lowest prediction error with the $\mathcal{Y}$, $\mathcal{P}$, and $\mathcal{S}$ as its input. The prediction error of using the depth information is higher than that of using the scene semantics as the depth is less discriminative than the scene segmentation as shown in Fig.~\ref{fig:tsne}. We show in Fig.~\ref{fig:net_structure_params_search} the accuracy of the CXA-Transformer with different structure settings, i.e., varying the number of Pre-LN layers, the dimension of the output of the embedding layers and that of the feed-forward sub-layers, and also the number of attention heads.

\subsection{Transfer to Robot Navigation}
To demonstrate that the egocentric trajectory forecasting ability learned from the sighted people navigating in the real world can be transferred to robots, enabling them to navigate in a socially compliant way, we conducted experiments on a real robot deployed in the crowd. The camera was mounted on top of the robot, which is about human chest height. Fig.~\ref{fig:robot_experiment_net_structure_params_search} shows two examples. In Fig.~\ref{fig:robot_experiment_1}, although there existed a chance for the robot to overtake the two people it was following, it still chose to keep following the people for a few seconds, as the model for predicting trajectories was learned from sighted people navigating in the real world (we used the one trained with $\mathcal{Y+P+S}$). As mentioned earlier, humans tend to follow certain social conventions while walking in the crowd. For example, in a scenario like the one shown in Fig~\ref{fig:robot_experiment_1}, instead of overtaking abruptly using the available space, which was in practice feasible, humans tended to keep following and not to disrupt the walking route of the people walking towards them. In this scenario, the robot well imitated such human behaviors and navigated in a socially polite way. In Fig~\ref{fig:robot_experiment_2}, the robot was moving in the opposite direction of the two people in its front. Our model predicted a safe trajectory for the robot to avoid collision with the people walking towards it.

\section{Discussion}
\label{sec:discussion}

With a person navigating in crowded spaces wearing a camera for data collection, the nearby people's navigation behaviors are less likely to be affected. Therefore, our dataset can be used to understand and model human navigation behaviours more precisely in egocentric settings compared to the data collected by a mobile robot. Our proposed end-to-end approach that utilizes multiple input modalities has also shown to be effective in forecasting the trajectory of the camera wearer. Although our experiments have shown that it is feasible to transfer human navigation behaviours to a mobile robot to enable socially compliant robot navigation, a successful transfer requires the robot to have a similar viewing angle and moving speed to the camera wearers who collected our dataset. Developing an effective adaptive method is needed especially if the robots have a different viewing angle. This is also applicable to assisting blind navigation, as a model learned from the sighted people data has to be adaptive to a different user, as different people have different heights, gaits, walking speeds, and reaction speeds. In addition, as our system directly forecasts the future trajectory that a blind person shall follow (this ability is learned from sighted people’s walking data), the system has to be initialized well (i.e., the blind person should be guided walking like a sighted person at the very first few minutes in order to allow the system to forecast meaningful trajectories). We leave implementing a complete blind navigation system as our future work. Integrating egocentric human trajectory forecasting with AR/VR applications is also practically useful. Future research can also be carried out in this direction.

\section{Conclusion}

In this work, we address the problem of forecasting the trajectory of camera wearers in the egocentric setting. To this end, a novel in-the-wild egocentric human trajectory forecasting dataset has been constructed, containing trajectories of the camera wearer in the 3D space and trajectories of nearby people in the 2D pixel space, as well as the segmented scene semantics and the estimated depth of the environment captured by the wearable camera. A transformer-based encoder-decoder model has been designed with a novel cascaded cross-attention mechanism to fuse features of different modalities for egocentric trajectory forecasting. We have also demonstrated that the trajectory forecasting ability learned from sighted people's navigation data can be transferred to a mobile robot, enabling socially compliant robot navigation.

\bibliographystyle{IEEEtran}


\end{document}